\documentclass[letterpaper, 10 pt, conference]{ieeeconf}  

\usepackage{graphicx}
\usepackage{amsmath}
\usepackage{amsfonts}
\usepackage{hyperref}
\hypersetup{
    colorlinks=true,
    linkcolor=black,
    citecolor=black,      
    urlcolor=black,
    }
\usepackage[dvipsnames]{xcolor}
\usepackage{dblfloatfix}
\usepackage{gensymb}
    
\IEEEoverridecommandlockouts                              

\title{\LARGE \bf
Multi-feature Compensatory Motion Analysis for Reaching Motions Over a Discretely Sampled Workspace
}

\author{Qihan Yang$^{1}$, \textit{Student Member}, IEEE, Yuri Gloumakov$^{2}$, Adam J. Spiers$^{3}$, \textit{Member}, IEEE
\thanks{$^{1}$Q. Yang (qihan.yang21@imperial.ac.uk) and $^{3}$A. J. Spiers (a.spiers@imperial.ac.uk) are with the Manipulation and Touch Lab (MTL),  Department of  Electrical and Electronic Engineering, Imperial College London, London, UK.}%
\thanks{$^{2}$Y. Gloumakov (yurigloum@berkeley.edu) is with the Mechanical Engineering Department, University of California, Berkeley, USA.}%
}
\begin{document}

\maketitle
\thispagestyle{empty}
\pagestyle{empty}

\begin{abstract}

The absence of functional arm joints, such as the wrist, in upper extremity prostheses leads to compensatory motions in the users' daily activities. Compensatory motions have been previously studied for varying task protocols and evaluation metrics. However, the movement targets' spatial locations in previous protocols were not standardised and incomparable between studies, and the evaluation metrics were rudimentary.
This work analysed compensatory motions in the final pose of subjects reaching across a discretely sampled 7$\times$7 2D grid of targets under unbraced (normative) and braced (compensatory) conditions. For the braced condition, a bracing system was applied to simulate a transradial prosthetic limb by restricting participants' wrist joints. A total of 1372 reaching poses were analysed, and a \textit{Compensation Index} was proposed to indicate the severity level of compensation. This index combined joint spatial location analysis, joint angle analysis, separability analysis, and machine learning (clustering) analysis. The individual analysis results and the final \textit{Compensation Index} were presented in heatmap format to correspond to the spatial layout of the workspace, revealing the spatial dependency of compensatory motions. The results indicate that compensatory motions occur mainly in a right trapezoid region in the upper left area and a vertical trapezoid region in the middle left area for right-handed subjects reaching horizontally and vertically. Such results might guide motion selection in clinical rehabilitation, occupational therapy, and prosthetic evaluation to help avoid residual limb pain and overuse syndromes.

\end{abstract}

\section{Introduction}

The two most common types of amputation in the upper arm are trans-humeral (between the shoulder and elbow joint, 28\%) and trans-radial (between the elbow and the wrist, 19\%)~\cite{amputation}. In both cases, an essential joint for manipulation is missing – the wrist. Amputees' lives have been made easier thanks to the flourishing of upper extremity prostheses (UEP). However, many commercially available UEP either do not include a functional wrist or only have a passive pronation and supination mechanism that must be rotated using another body part. Therefore, when such UEP users try to perform reach and grasp or pick and place tasks, their prosthetic forearm is locked into a single orientation~\cite{Bajaj2015}. This lack of mobility in the arm leads to compensatory motions.

Compensatory motions are abnormal arm or body movements to compensate for lacking Degrees of Freedom (DoF) in arm joints, often resulting in residual limb pain and overuse syndromes~\cite{{Montagnani2015}}, and therefore, have been widely studied for decades. In \cite{Hussaini2017}, compensatory motions were detected in six clinical motions (hanging cloth, slicing, stirring, sweeping, eating, and cutting) performed by four prosthesis users. In \cite{Kato2018}, the captured movements were a series of pick-and-place tasks. Recently, more research has focused on the heterogeneous characteristics of different users. In \cite{Engdahl2022} and \cite{Valevicius2020}, individual differences in compensatory motions were studied, where prosthesis users were asked to perform both pick-and-place and manipulation tasks. The results showed that the prosthesis users' skills and experience affect their moving strategies and performances. 

In many compensatory motion studies that were not able to find UEP user participants, bypass sockets - a mechanical interface that mounts a terminal device to the forearm of a non-disabled person to achieve a reasonable simulation of prosthetic use~\cite{Williams2021} - are commonly used. In \cite{Carey2008} and \cite{Paskett2019}, movements such as drinking from a cup, opening a door, and modifying objects were performed by non-disabled participants with and without by-pass sockets and compensatory motions were observed between the two conditions.

\begin{figure}[t]
      \centering
      \includegraphics[width=0.8\linewidth]{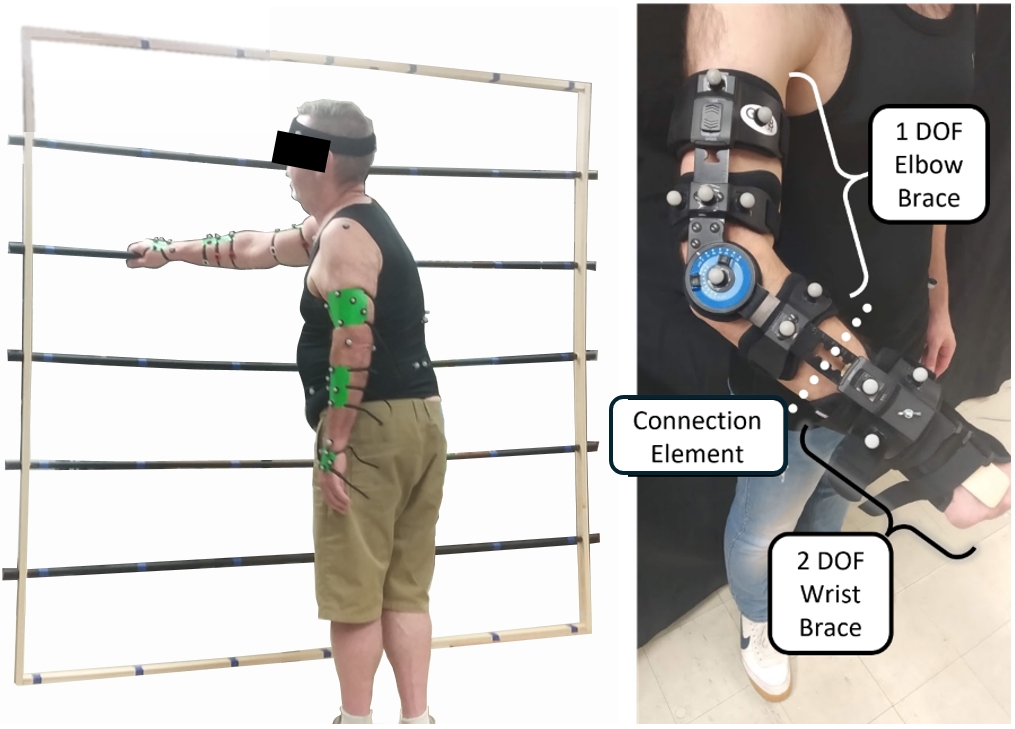}
      \caption{Experiment setup (horizontal placement) and the custom arm bracing system.}
      \label{Fig: setup}
      \vspace{-10pt}
\end{figure}

\begin{figure*}[t]
      \centering
      \includegraphics[width=\linewidth]{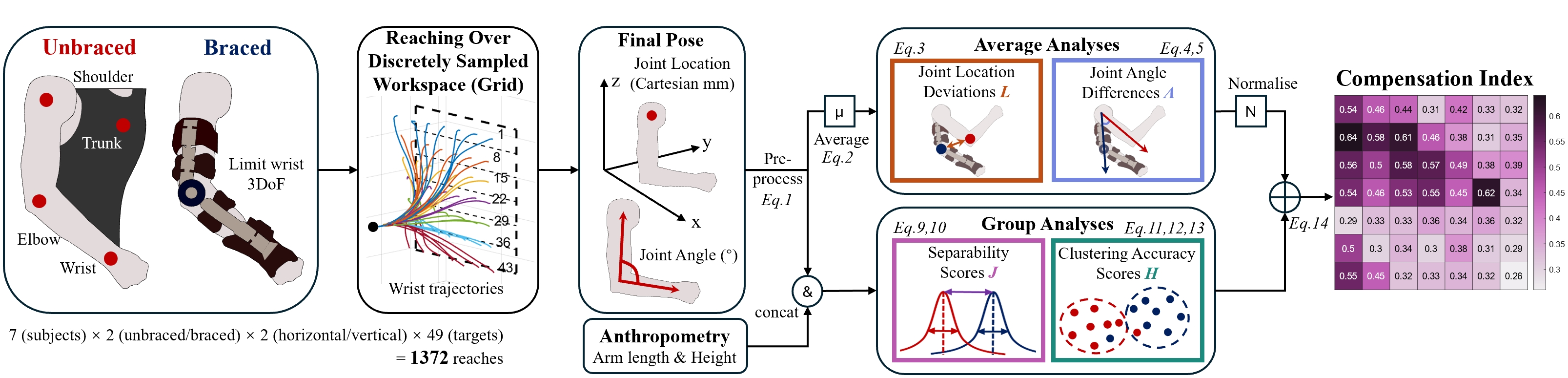}
      \caption{The workflow of this project. A total of 1372 reaching data are collected from 7 subjects under the unbraced and braced condition, reaching the 49 targets horizontally or vertically. The reaching final pose (final joint locations and angles) and subjects' static anthropometry information are used to calculate four compensatory motion evaluation metrics: average joint location deviation, average joint angle difference, group separability score, and group clustering accuracy score. The four components are combined as the \textit{Compensation Index} and presented in heatmap format.}
      \label{Fig: workflow}
      \vspace{-6pt}
\end{figure*}

Though compensatory motions have been studied using various movements, they were designed based on some subset of Activities of Daily Living (ADL) tasks, and their targets' spatial locations were not standardised. 
This limits the possibility of forming a holistic view of how compensatory motion happens across the human-reaching workspace. One study standardised the movements using the Southampton Hand Assessment Protocol (SHAP)~\cite{Light2002}. Murgia et al. monitored and evaluated the compensatory motions in the SHAP page-turning task~\cite{Murgia2010}. SHAP provides a standard hand/arm study protocol, using various objects and actions, but all the manipulations are limited to occurring on a desk. In another approach to address the broader workspace characterisation, Chadwell et al. asked subjects to sweep their hand through nine arcs in the frontal, transverse, and sagittal planes with the elbow fully extended~\cite{Chadwell2020}. However, the subject trunk was limited to an upright posture, eliminating the spine's rotation and lateral flexion abilities.

Besides problems in movements, previous compensatory motion evaluation metrics were monotonous. A substantial amount of previous research only investigated compensatory motion on a single joint (either shoulder or trunk), with evaluation metrics limited to the average joint Range of Motion (ROM)~\cite{Hussaini2017, Engdahl2022, Valevicius2020, Carey2008, Paskett2019, Murgia2010}. A diverse metric considering more joints and evaluation aspects is needed.

Our work differs from previous efforts in analysing compensatory motions in the final pose of subjects reaching across a discretely sampled workspace under unbraced (natural) and braced (compensatory) conditions. A custom arm bracing system that restricts wrist movement in 3DoF was used in the braced condition to simulate a bypass socket or prosthetic arm without an active wrist (Fig. \ref{Fig: setup}). A novel \textit{Compensation Index} was proposed to evaluate the compensatory motion severity level between the unbraced and braced conditions. This index considered the compensatory motions in the elbow, shoulder, and trunk. It was estimated from four perspectives: joint location analysis, joint angle analysis, separability analysis, and machine learning (clustering) analysis. The results were presented in heatmap format to correspond to the workspace's spatial layout, revealing the compensatory motions' spatial dependency. This insight gives guidance on motion selection in clinical rehabilitation, occupational therapy, and prosthetic evaluation, avoiding residual limb pain and overuse syndromes. The workflow of this project is shown in Fig. \ref{Fig: workflow}.

\section{Dataset}

\begin{figure*}[t]
      \centering
      \includegraphics[width=\linewidth]{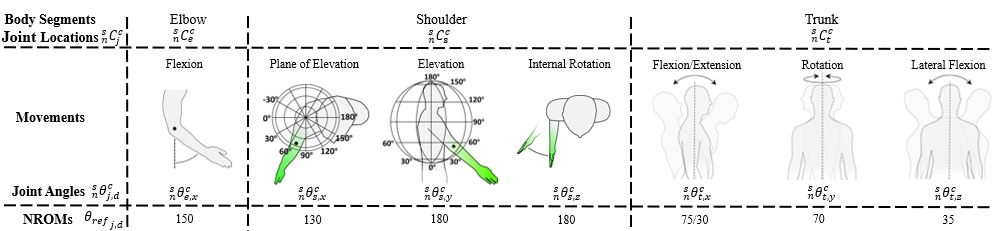}
      \caption{The first row includes the nomenclatures of the body segment and the symbolic representations of joint locations. The second and third rows include the nomenclatures, graphic and symbolic representations and Normal Range of Motion (NROM) of the joint movement along each orientation.}
      \label{Fig: joint movements}
      \vspace{-5pt}
\end{figure*}

The dataset used in this study is from \cite{Spiers2024}, in which the workspace in front of the subjects was discretely sampled by a 7$\times$7 square grid at 300mm intervals, horizontally and vertically. The subjects - seven non-disabled, right-handed adults – were asked to reach to and grasp each of the 49 cylindrical targets (one at a time) either orientated horizontally or vertically, depending on the placement of the grid. Each reaching motion started with the user's arm hanging at their side and ended when they grasped the target. A sample horizontal reaching configuration is shown in the left part of Fig.\ref{Fig: setup}. The 49 reaches were performed twice under two conditions: unbraced and braced. Under the braced condition, subjects wore a combination of 1DOF elbow brace and 2DOF wrist brace, as shown in the right part of Fig. \ref{Fig: setup}. The combined brace limited 3DoF of the wrist: flexion and extension, radial and ulnar deviation, and pronation and supination. The reaching motions were recorded using a motion capture system consisting of 12 Vicon Bonita cameras. The markers were attached to the subject's pelvis, thorax, shoulder, elbow, and wrist, from which the joint angles of 7 movements (elbow flexion, shoulder plane of elevation, elevation and internal rotation, trunk flexion/extension, rotation and lateral flexion) were calculated (graphic representations are in Fig. \ref{Fig: joint movements}). 

In summary, the dataset contains 7 (subjects) $\times$ 2 (braced / unbraced) $\times$ 2 (horizontal / vertical) $\times$ 49  (targets) = 1372 reaches. The reaching intervals, joint xyz locations in Cartesian space (mm) and joint angles (degree) during reaches, and participants' information (age, height, arm length, etc) are available. Details about the dataset and study protocol are available in~\cite{Spiers2024}, and \url{https://www.imperial.ac.uk/manipulation-touch/open-source/data/}.

\section{Method}
Our work differs from previous studies and \cite{Spiers2024} in focusing on the final pose of reaching motions. The final pose is the elbow, shoulder, and trunk locations and corresponding joint angles when the subjects finish reaching and grasping the target (the final time step in the reaching intervals). 
For each subject $s$, the final elbow, shoulder, and trunk ($j\in\{e,s,t\}$) locations in Cartesian space and the 7 joint angles along the corresponding orientation ($d\in\{x,y,z\}$) when reaching target $n$ under unbraced and braced conditions ($c\in\{u,b\}$) were extracted and denoted as ${}_n^sC_{fin}{}_j^c\in \mathbb{R}^{1\times3}$ and ${}_n^s\theta_{fin}{}_{j,d}^c\in \mathbb{R}$, respectively.


Problems remain in ${}_n^sC_{fin}{}_j^c$ and ${}_n^s\theta_{fin}{}_{j,d}^c$. The motion capture system origin of subject 3 was set differently than others, making the absolute final joint location incomparable between subjects. Moreover, some joint angles outweigh others (e.g., shoulder elevation spans from 34 to 130 degrees, larger than trunk lateral flexion, which spans from -12 to 15 degrees). The final joint locations and angles were preprocessed to eliminate the errors introduced in data collection and make the joint angles in a comparable range (\ref{eq: norm_loc_angle}). For subject $s$ reaching under condition $c$, an identical reference - the average of the 49 initial joint locations (${}_n^sC_{ini}{}_j^c$) - was subtracted from the 49 final joint locations. This transferred the absolute final joint location to relative joint translation, making the resulting joint locations (${}_n^sC_j^c$) only dependent on the relative position of the subjects and the grid (consistent for all cases) but not the motion capture system origin. Following clinical conventions~\cite{rom}, the final joint angle along each orientation was normalised by the corresponding Normal Range of Motion (NROM) ($\theta_{ref}{}_{j,d}$ in Fig. \ref{Fig: joint movements}). The resulting relative joint angles (${}_n^s\theta_{j,d}^c$) have a comparable range from 0 to 70.
\begin{equation}
\label{eq: norm_loc_angle}
    {}_n^sC_j^c={}_n^sC_{fin}{}_j^c-\frac{1}{49} \sum_{n=1}^{49}{}_n^sC_{ini}{}_j^c
    \quad \Bigg| \quad
    {}_n^s\theta_{j,d}^c=\frac{{}_n^s\theta_{fin}{}_{j,d}^c}{\theta_{ref}{}_{j,d}} \times 100 
\end{equation}

Compensatory motion evaluation metrics measure whether or to what extent subjects act differently after the arm brace is applied. Our metric, the \textit{Compensation Index}, improved upon previous ROM metrics, evaluates compensatory motion from four aspects: 
\begin{enumerate}
    \item Average joint location deviation, $L$,
    \item Average joint angle difference, $A$,
    \item Group separability score, $J$,
    \item Group clustering accuracy score, $H$,
\end{enumerate}

The average joint location deviation and angle difference analyses evaluate compensatory motions from the subjects' average performance. The separability and clustering analyses take individual differences into account. The two analytical subgroups will now be further described. 

\subsection{Average Performance Analyses}
The average performance analyses align with~\cite{Spiers2024}, reflecting, on average, how differently the subjects perform after the arm brace is applied when reaching a target in space.
\subsubsection{Average Joint Location and Angle}
Given $c$, $j$, and $n$, the average joint locations ($\overline{{}_nC_j^c}$) and average joint angles ($\overline{{}_n\theta_{j,d}^c}$) of the seven subjects were calculated as (\ref{eq: ave}).
\begin{equation}
\label{eq: ave}
    \overline{{}_nC_j^c}= \frac{1}{7}\sum_{s=1}^7{}_n^s{C_{j}^c}
    \quad \Bigg| \quad
    \overline{{}_n\theta_{j,d}^c} = \frac{1}{7}\sum_{s=1}^7{}_n^s{\theta_{j,d}^c} \qquad 
\end{equation}

\subsubsection{Joint Location Deviation}
At each target $n$, the average joint location deviation (Euclidean distance) between the unbraced ($c=u$) and braced ($c=b$) conditions was calculated for each joint centre (${}_nL_{e/s/t}$). The final joint location deviation at target ${}_nL$ was the unweighted average (the three joints are assigned with equal importance) of the three individual joint location deviations (\ref{eq: loc_div}). 
\begin{equation}
\label{eq: loc_div}
    {}_nL_j=\left\|\overline{{}_nC_j^b}-\overline{{}_nC_j^u}\right\|
    \rightarrow
    {}_nL = \frac{1}{3}\left( {}_nL_e+{}_nL_s+{}_nL_t\right)
\end{equation}


\subsubsection{Joint Angle Difference}


At each target $n$, the absolute average joint angle difference between the unbraced and braced conditions was calculated for each joint (${}_nA_{e/s/t}$) (\ref{eq: angle_diff_joint}). The elbow angle differences (${}_nA_e$) were solely from elbow flexion ($\overline{{}_n\theta_{e,x}^c}$). The shoulder and trunk angle differences (${}_nA_{s/t}$) were the unweighted average of the three angle differences along the three orientations ($\overline{{}_n\theta_{s/t,x/y/z}^c}$). The final average joint angle difference ${}_nA$ was the unweighted average of the three individual joint angle differences (\ref{eq: angle_diff}). 
\begin{equation}
\label{eq: angle_diff_joint}
    {}_nA_e\!=|\overline{{}_n\theta_{e,x}^b}\!-\overline{{}_n\theta_{e,x}^u}|
    \text{ }\Bigg|\text{ }
    {}_nA_{s/t}\! = \frac{1}{3}\sum_{d}|\overline{{}_n\theta_{s/t,d}^b}\!-\overline{{}_n\theta_{s/t,d}^u}|
\end{equation}
\begin{equation}
\label{eq: angle_diff}
    {}_nA = \frac{1}{3}\left({}_nA_e+ {}_nA_s+ {}_nA_t\right)
\end{equation}

Large average performance discrepancy (${}_nL$, ${}_nA$) implies that, on average, greater compensatory motions occur at target location $n$. However, the average performance is insufficient to reveal the severity level of compensatory motions without studying the consistency among subjects.

\subsection{Group Performance Analyses}
\label{sec: diff_analysis}
It was hypothesised that even if a target has a large average performance discrepancy, the individual difference at that target can be remarkable (different subjects take different reaching strategies), making the joint locations/angles data from all subjects dispersed and highly overlap between the unbraced and braced conditions. If so, the average performance cannot fully represent the common characteristic of compensatory motions (it is inappropriate to claim more compensatory motions occur at a target only based on a large average performance discrepancy). To validate such hypothesis, inspired by \cite{Engdahl2022, Valevicius2020}, the joint location and angle standard deviation (${}_n\sigma_C^c, {}_n\sigma_\theta^c$) at each target $n$ were calculated for both unbraced or braced conditions (\ref{eq: std_angle},\ref{eq: std_loc}). 

\begin{equation}
\label{eq: std_loc}
    {}_n\sigma_C^c=\frac{1}{3}\left(\operatorname{std}({}_nC_e^c)+\operatorname{std}({}_nC_s^c)+\operatorname{std}({}_nC_t^c)\right)
\end{equation}
\begin{equation}
\label{eq: std_angle}
    {}_n\sigma_\theta^c=\frac{1}{3}\left[\operatorname{std}({}_n\theta_{e,x}^c)+\frac{\sum_d \operatorname{std}({}_n\theta_{s,d}^c)}{3} +\frac{\sum_d \operatorname{std}({}_n\theta_{t,d}^c)}{3} \right]
\end{equation}

The group performance analyses, utilising a feature vector, take the individual differences into account, reflecting the similarity between the features from the unbraced and the braced conditions.

\subsubsection{Feature Vector}
The feature vector ${}_n^sx_j^c\in \mathbb{R}^{1 \times 6|8}=\left[{}_n^sC_j^c, {}_n^s\theta_{j,\forall d}^c, {}^sh, {}^sl\right]$ combined the joint location, joint angle, and subjects' static anthropometry information (height ${}^sh$ and arm length ${}^sl$). The feature matrix of the seven subjects was denoted as ${}_nX_j^c \in \mathbb{R}^{7 \times 6|8}$. Given $j$ and $n$, the average feature of the seven subjects under each condition ($\overline{{}_nx_j^c}$) and both conditions ($\overline{{}_nx_j}$) were calculated as (\ref{eq: ave_all}).
\begin{equation}
\label{eq: ave_all}
\overline{{}_nx_j^c}= \frac{1}{7}\sum_{s=1}^7{}_n^sx_j^c
    \quad \Bigg| \quad
    \overline{{}_nx_j} = \frac{1}{2}\left(\overline{{}_nx_j^u}+\overline{{}_nx_j^b}\right)
\end{equation}

\subsubsection{Separability Score}
Inspired by~\cite{Felipe2022}, at each target $n$, the within-class variance (${}_nS_{Wj}$) and between-class variance (${}_nS_{Bj}$) of the features from unbraced and braced conditions were calculated for each joint. The joint separability score (${}_nJ_{e/s/t}$) was the ratio of the two variances (\ref{eq: sep_joint}). The final separability score ${}_nJ$ was the unweighted average of the three individual joint separability scores (\ref{eq: sep}).
\begin{equation}
\label{eq: sep_joint}
\left.\begin{array}{c}
{}_nS_{W, j}=\sum_{c} \sum_s\left\|{}_n^s x_j^c-\overline{{}_nx_j^c}\right\|^2 \\
{}_nS_{B, j}=\sum_{c} 7\times\left\|\overline{{}_nx_j^c}-\overline{{}_nx_j}\right\|^2
\end{array}\right\} {}_nJ_j=\frac{{}_nS_{B, j}}{{}_nS_{W, j}}
\vspace{-5pt}
\end{equation}
\begin{equation}
\label{eq: sep}
    {}_nJ=\frac{1}{3} \left( {}_nJ_e+ {}_nJ_s+ {}_nJ_t\right)
\end{equation}


\subsubsection{Clustering Accuracy Score}
Separability analysis requires less computational cost, but outliers can corrupt the result~\cite{Felipe2022}. A more robust method is the clustering analysis. At each target $n$, joint $j$'s feature matrices from the unbraced (${}_nX_j^u$) and braced (${}_nX_j^b$) conditions were combined as input and separated into two clusters (${}_nK_j^1, {}_nK_j^2$) based on a clustering algorithm (\ref{eq: cluster_joint}). 
\begin{equation}
\label{eq: cluster_joint}
    \left({}_nK_j^1, {}_nK_j^2\right)=\operatorname{ Clustering }\left([{}_nX_j^u, {}_nX_j^b]\right)
\end{equation}

The best match between the generated clusters and the actual labels was found (if cluster 1 contains more data from the unbraced condition, match cluster 1 to the unbraced condition and cluster 2 to the braced condition). The joint clustering accuracy (${}_nH_{e/s/t}$) was calculated as the number of correctly clustered data from the two matches (${}_nN_j^{\operatorname{true}}$) over the total amount of data ($N=14$) (\ref{eq: match}). The final clustering accuracy score ${}_nH$ was the unweighted average of the three individual joint clustering accuracies (\ref{eq: cluster}).
\begin{equation}
\label{eq: match}
{}_nN_j^{\operatorname{true}}=\operatorname{Match}\left({}_nX_j^u, {}_nX_j^b, {}_nK_j^1, {}_nK_j^2\right)
\vspace{-3pt}
\end{equation}
\begin{equation}
\label{eq: cluster}
{}_nH_j ={}_nN_j^{\operatorname{true}}/{N}
\rightarrow
    {}_nH=\frac{1}{3} \left( {}_nH_e+ {}_nH_s+ {}_nH_t\right)
\end{equation}

In this work, the clustering algorithm was Agglomerative Hierarchical Clustering. Two distance metrics (Manhattan and Euclidean) and three linkages (complete, average, and single) were implemented. At each target, the combination that gave the highest clustering accuracy score was adopted. 

Large group performance discrepancy (${}_nJ$, ${}_nH$) suggests the differences between the collective performance of the seven subjects reaching with and without the arm braced are more distinct at target $n$. It can be reasonably concluded that compensatory motions occur more commonly at a target location if it has both large average and group performance discrepancies (large joint location deviations and angle differences, and the subjects reaching strategies are consistent). 

\subsection{Compensation Index}
The proposed \textit{Compensation Index}, as a diverse multi-modal compensatory motion evaluation metric, reflected the compensatory motion severity level at a target location by combining the four metrics (${}_nL, {}_nA, {}_nJ, {}_nH$). 

The first step was to make the four components comparable in range. Regular normalisation methods, such as min-max normalisation, are not suitable. Such a normalisation method would set the maximum value in this research to one, but another experiment with different users/conditions might observe more significant compensatory motions and would exceed that bound of normalisation. Hence, to make the four components comparable and able to be adapted to other research, they were normalised to the same level using different approaches empirically: ${}_nL$ was divided by 100 (equivalent to using `meter' as the unit), ${}_nA$ was divided by 10, ${}_nJ$ and ${}_nH$ remained unchanged. After this empirical normalisation, the four components were comparable in range, and new observations from other research could be added to the analysis following the same normalisation. The final \textit{Compensation Index} ${}_nI$ took the unweighted average of the four normalised components (\ref{eq: index}).
\begin{equation}
\label{eq: index}
    {}_nI = \frac{1}{4}\left(\frac{1}{100}{}_nL+\frac{1}{10}{}_nA+{}_nJ+{}_nH \right)
\end{equation}


\subsection{Heatmap Representation}
\begin{figure}[b]
\vspace{-10pt}
      \centering
      \includegraphics[width=0.99\linewidth]{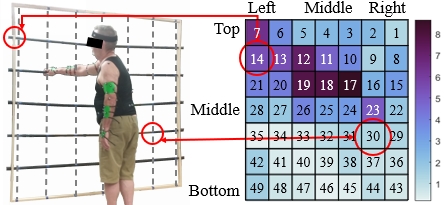}
      \caption{Heatmap representation and the relationship with the grid. The number in each grid indicates the number of the target location.}
      \label{Fig: heatmap}
\end{figure}

\setcounter{figure}{5}
\begin{figure*}[b]
\vspace{-12pt}
      \centering
      \includegraphics[width=1\linewidth]{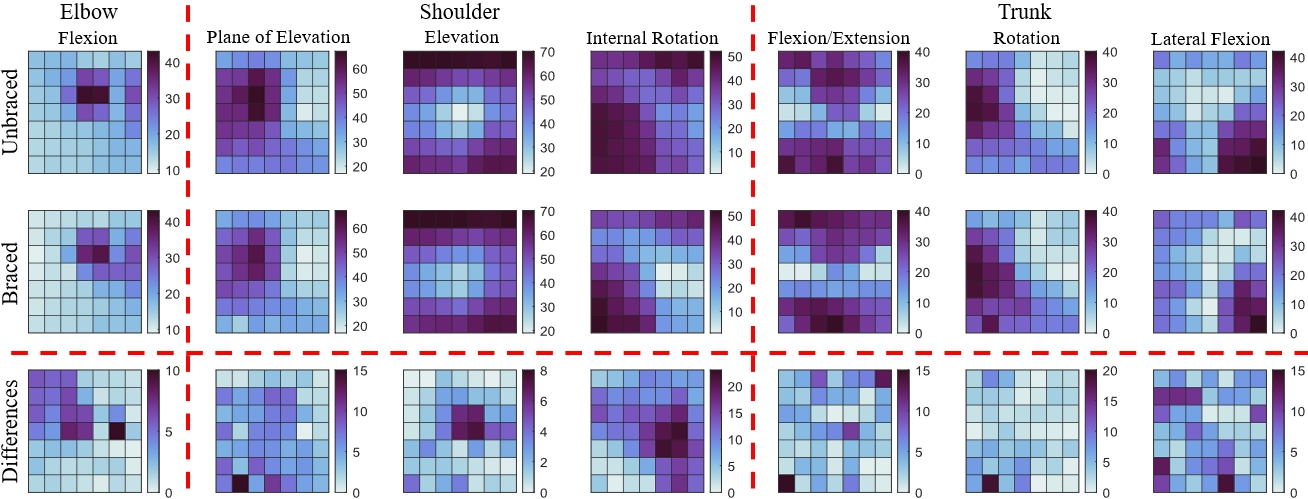}
      \caption{Each joint movement's average joint angles under the unbraced and braced conditions, and the absolute differences between them.}
      \label{Fig: joint angles and difference}
\end{figure*}

This work is particularly interested in whether the compensatory motions depend on the target spatial location. Inspired by \cite{Spiers2018,Spiers2024}, a heatmap representation was implemented as shown in Fig. \ref{Fig: heatmap}. Each position in the 7$\times$7 heatmap corresponds to a target location in the 7$\times$7 grid, seeing from the subjects' perspective. The deeper colour represents the larger value of ${}_nL$, ${}_nA$, ${}_nJ$, ${}_nH$, or ${}_nI$.

\section{Results and Discussion}
The following sections present the detailed results of the four metrics when the subjects grasp horizontally and the \textit{Compensation Index} for both horizontal and vertical targets.

\subsection{Average Performance Analyses Results}
\subsubsection{Joint Location Deviation}

\setcounter{figure}{4}
\begin{figure}[b]
\vspace{-10pt}
      \centering
      \includegraphics[width=0.97\linewidth]{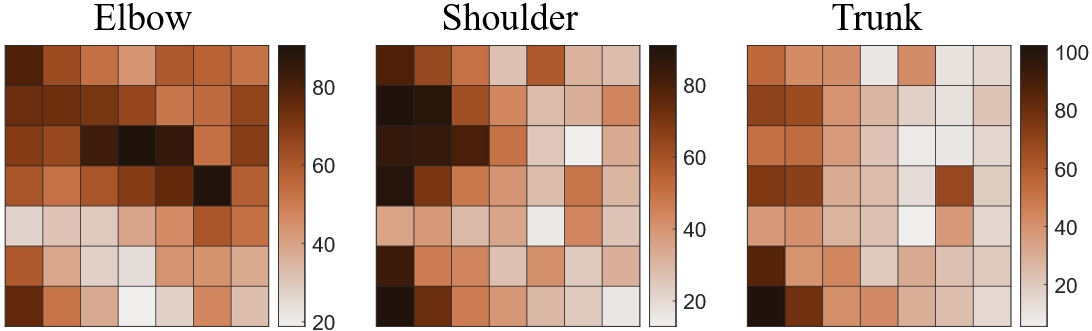}
      \caption{Elbow, shoulder, and trunk average joint location deviations.}
      \label{Fig: location deviations}
\end{figure}

Elbow, shoulder, and trunk joint location deviations between the two conditions at the 49 targets are shown in Fig. \ref{Fig: location deviations}. The elbow has the most significant deviation (the average of the 49 targets is 55.1). A reasonable explanation is that the elbow is closer to the extremity, and the deviations in the trunk and shoulder transmit to the deviations in the elbow. All three individual joint location deviations are distinctly spacial dependent: elbow deviations are majorly located at the centre and upper left area; shoulder deviations are majorly located at the upper left area; all three joints have deviations at the bottom left corner. The final average joint location deviations ($L$) are shown in the left part of Fig. \ref{Fig: average results}, and the significant deviations are in the upper left area, bottom left corner, and target 23. 


\subsubsection{Joint Angle Difference}

The seven average joint angles under the unbraced and braced conditions and the differences between them are shown in Fig. \ref{Fig: joint angles and difference}. The joint movement corresponding to the largest angle differences is shoulder internal rotation (the average of the 49 targets is 10.6). It also shows a strong spatial dependency with most differences in the lower right area. Other spatial-dependent joint movements are elbow flexion and shoulder elevation. For elbow flexion, the differences are located at the upper left area and target 23. For shoulder elevation, the differences are located at the centre area. The final average joint angle differences ($A$) are shown in the right part of Fig. \ref{Fig: average results}, where the significant differences are in the centre area and the bottom left corner. 


According to the average analyses, the upper left, bottom left, and centre areas observe more compensatory motions in subjects' average performance after applying the arm brace. These results, aligned with~\cite{Spiers2024} but more integrated, emphasise the spatial dependency of the compensatory motions and form the cornerstone to understanding the compensatory motions' severity level across the human-reaching workspace. The next step is to integrate the results with individual differences.

\setcounter{figure}{6}
\begin{figure}[t]
      \centering
      \includegraphics[width=0.95\linewidth]{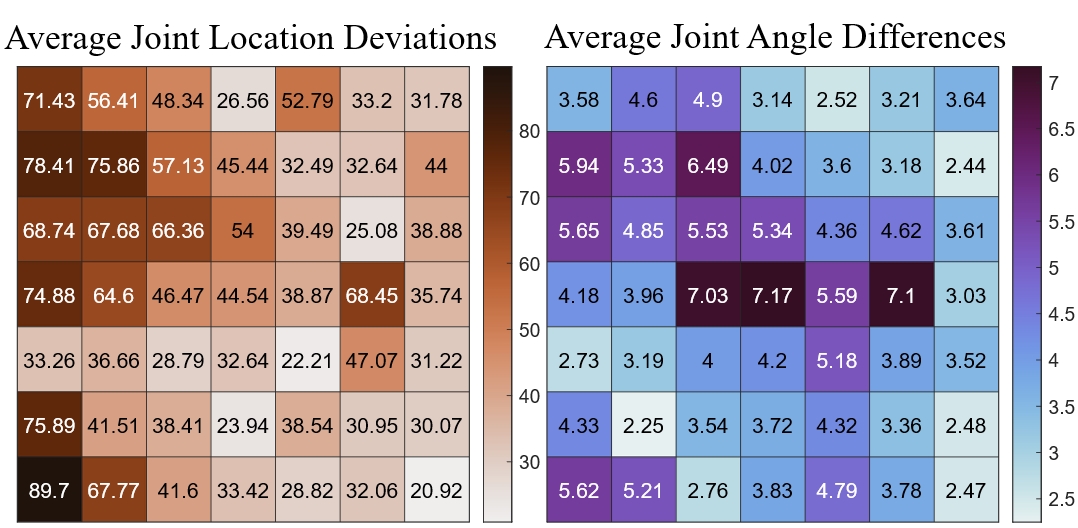}
      \caption{Average joint angle differences (left) and location deviations (right).}
      \label{Fig: average results}
      \vspace{-16pt}
\end{figure}
\begin{figure}[b]
\vspace{-10pt} 
      \centering
      \includegraphics[width=\linewidth]{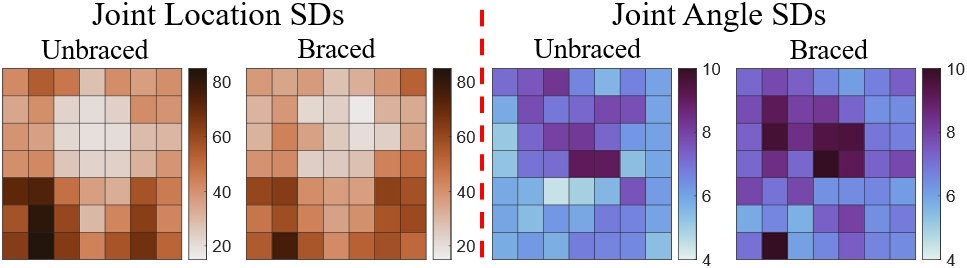}
      \caption{Joint angle and location standard deviations of the two conditions.}
      \label{Fig: std}
\end{figure}

\subsection{Individual Difference Analysis Results}
\label{Sec: variation}
According to Fig. \ref{Fig: std}, some targets (e.g. $n=5, 6, 20, 48$) not only have large average joint discrepancies (${}_nL,{}_nA$) but also have large individual differences (${}_n\sigma_C^c,{}_n\sigma_\theta^c$). This indicates that the subjects take different reaching strategies at those target locations, which, as discussed in section \ref{sec: diff_analysis}, reduce the significance of the average performance analyses. The individual difference analysis results validate the importance of the group performance analyses.


\subsection{Group Performance Analyses Results}
\begin{figure}[b]
    \vspace{-10pt}
      \centering
      \includegraphics[width=0.97\linewidth]{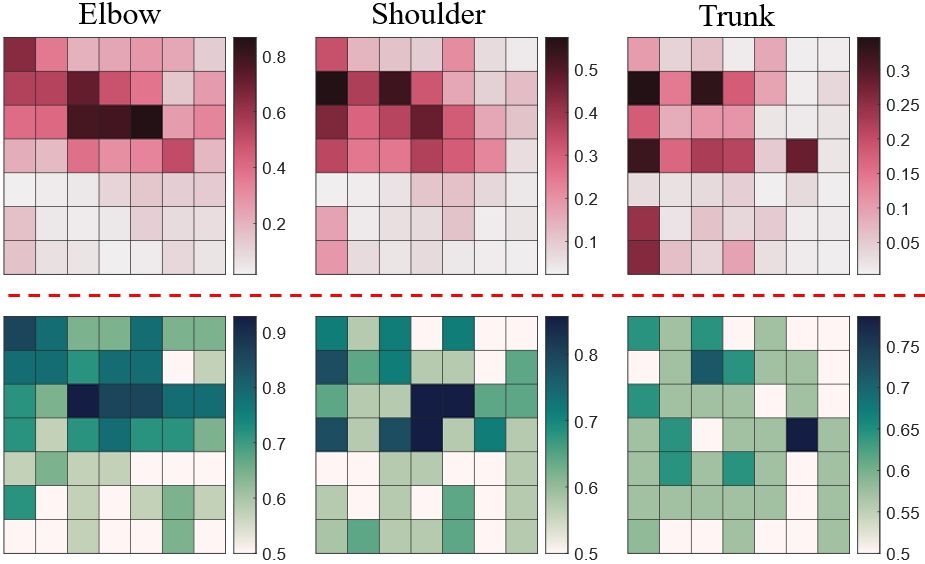}
      \caption{Elbow, shoulder, and trunk separability scores (upper) and clustering accuracy scores (lower).}
      \label{Fig: sep and cluster}
\end{figure}
\begin{figure}[t]
      \centering
      \includegraphics[width=0.95\linewidth]{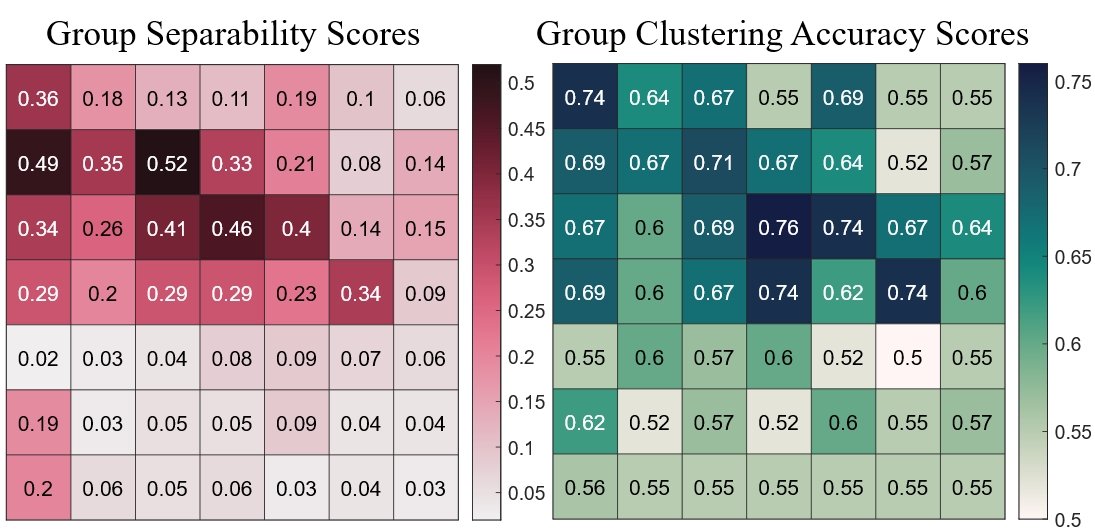}
      \caption{Separability (left) and clustering accuracy scores (right).}
      \label{Fig: group results}
      \vspace{-16pt}
\end{figure}

\subsubsection{Separability Score}
Elbow, shoulder, and trunk separability scores are shown in the upper part of Fig. \ref{Fig: sep and cluster}. The elbow has the largest separability score (the average of the 49 targets is 0.26). All three individual joint separability scores are spatial dependent with larger separations at the upper left area. The trunk has outliers at the bottom left. The final separability scores are shown in the left part of Fig. \ref{Fig: group results}, where the large values are in the upper left area. 

\subsubsection{Clustering Accuracy Score}

Elbow, shoulder, and trunk clustering accuracy scores are shown in the lower part of Fig.\ref{Fig: sep and cluster}. The elbow has the highest average clustering accuracy (the average of the 49 targets is 0.65). The elbow and shoulder clustering accuracy scores are spatial dependent: for the elbow, the high values are located in the upper left area; for the shoulder, the high values are located in the centre area. The final clustering accuracy scores are shown in the right part of Fig. \ref{Fig: group results}, where the large values are in the centre and upper left area. 

According to the group analyses, the collective performance of the seven subjects reaching with and without the arm braced are more distinct at the centre and upper left area. This finding is crucial supplementary information to add to the average performance analyses.

\subsection{Compensation Index}
\begin{figure}[b]
\vspace{-10pt}
      \centering
      \includegraphics[width=\linewidth]{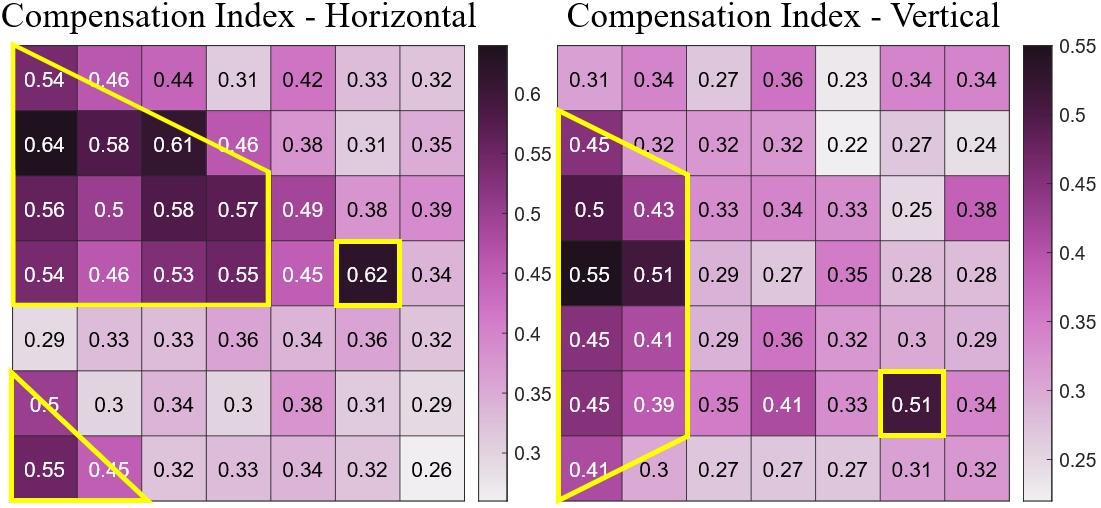}
      \caption{Compensation index for the subjects reaching horizontally (left) and vertically (right), with large values highlighted in yellow lines.}
      \label{Fig: index}    
\end{figure}

The compensation indexes for the horizontal and vertical targets are shown in Fig. \ref{Fig: index}. For the horizontal target, the large values are mostly in a right trapezoid region at the upper left area, with some outliers at the bottom-left corner. Note that the compensation index at target 23 is outstandingly large compared to the values around it.
For the vertical target, the large values are mostly in a vertical trapezoid region at the left two columns of the grid. Similar to the horizontal result, there is a significant value at target 37 in the same column as target 23 but two rows below. Given the comparability between the horizontal and vertical configurations, having these outstanding values in the right area of the workspace should not be caused by error but by distinct reaching strategies.

These results reveal the common severity level and the spatial dependency of compensatory motions across the human-reaching workspace. When a functional wrist is absent, the compensatory motions are more severe when subjects reach horizontally to the upper left and vertically to the middle left area of the workspace. When reaching the right area of the workspace, theoretically more common for right-handed subjects, the compensatory motion is milder. However, significant compensatory motions occur when subjects reach horizontally to a target around the right thorax and waist. Similar affection appears for vertical reaches once the target location moves down to the right knee and shin. 

\begin{figure*}[t]
      \centering
      \includegraphics[width=0.6\linewidth]{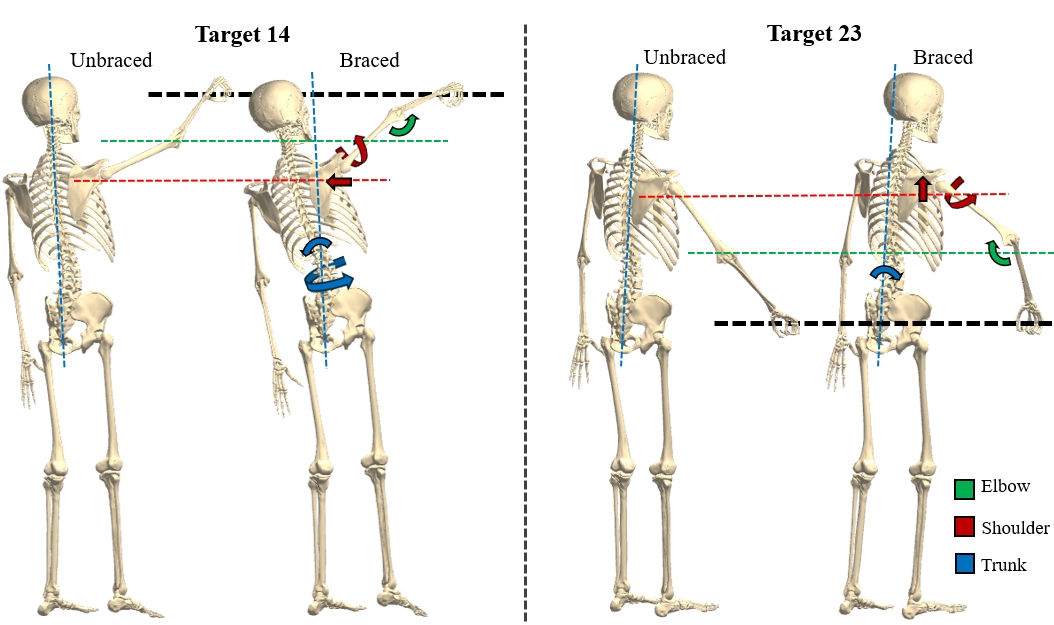}
      \caption{Reconstructed average final poses of subjects reaching horizontally to target 14 and 23 under the unbraced and braced conditions.}
      \label{Fig: reconstruction}
      \vspace{-12pt}
\end{figure*}

To visualise the compensatory motions, the average final poses of the seven subjects grasping targets 14 and 23 horizontally, under the unbraced and braced conditions, are reconstructed using KineBody 3D movable human skeleton model, as shown in Figure.\ref{Fig: reconstruction}. 

\section{Conclusion}


This work quantitatively analysed the compensatory motions in the final pose of subjects reaching across a 7$\times$7 discretely sampled workspace under unbraced and braced conditions. A \textit{Compensation Index}, combining the average joint location deviation, average joint angle difference, group separability score, and group clustering accuracy score in the elbow, shoulder, and trunk, was proposed to indicate the common severity level of compensation. The results, presented in heatmap format, revealed the spatial dependency of compensatory motions: for right-handed subjects, the absence of a functional wrist significantly affects reaching horizontally to the upper left area and vertically to the middle left area of a workspace in front of subjects. When designing clinical rehabilitation / occupational therapy approaches or designing and evaluating prosthetics for right-handed subjects, movements that involve reaching the mentioned area should be carefully examined and, in general, avoided in case of residual limb pain and overuse syndrome. 

There are limitations in this work. The three joints were assigned and averaged with equal weight when calculating the four individual metrics. Similarly, the four individual metrics were assigned and averaged with equal weight when calculating the final Compensation Index. These conventions could be modified under reliable clinical suggestions. Moreover, the dataset could be optimised. One fundamental improvement can be achieved by introducing more repetition at each location. Another limitation is the grid does not exhaustively explore the upper arm functions (only power grasp was tested, and the target is a weightless tube). Future studies are encouraged to explore more grasp types and take objects' weight and shape into account. Moreover. the data were collected in a lab environment, which could make participants nervous and act unnaturally, causing bias in the results. A foreseeable next step is to transfer the study to upper-limb amputees in a home environment using wearable motion capture sensor systems.


\bibliographystyle{IEEEtran}
\bibliography{ref}

\end{document}